\documentclass[11pt]{article}

\usepackage[final]{acl}

\usepackage{times}
\usepackage{latexsym}

\usepackage[T1]{fontenc}

\usepackage[utf8]{inputenc}

\usepackage{microtype}

\usepackage{inconsolata}

\usepackage{graphicx}

\usepackage{float} 
\usepackage{url}
\usepackage{verbatim}
\usepackage{graphicx}
\usepackage{caption}
\usepackage[labelfont=bf]{caption}
\usepackage{multirow}
\usepackage{booktabs}
\usepackage{adjustbox}
\usepackage{pifont}
\usepackage{enumitem}
\usepackage{amsmath} 
\usepackage{amssymb} 

\usepackage{amssymb}  
\usepackage{booktabs} 
\usepackage{array}    

\usepackage[ruled,vlined]{algorithm2e}
\SetAlgoSkip{small} 
\SetAlgoNoLine 
\DontPrintSemicolon 
\SetAlCapHSkip{0pt} 
\SetAlgoNoEnd 
\SetAlgoVlined 
\SetAlgoCaptionSeparator{.\ } 
\SetAlFnt{\small} 

\usepackage{tabularx} 
\usepackage{booktabs} 
\usepackage{multirow} 

\usepackage{makecell}

\usepackage[most]{tcolorbox}

\newtcblisting{promptbox}{
  colback=gray!5,
  colframe=black!40,
  boxrule=0.5pt,
  arc=2pt,
  left=6pt,
  right=6pt,
  top=6pt,
  bottom=6pt,
  listing only,
  listing options={
    basicstyle=\ttfamily\small,
    columns=fullflexible,
    keepspaces=true,
    breaklines=true
  },
  breakable
}

\title{ZoFia: Zero-Shot Fake News Detection with Entity-Guided Retrieval and Multi-LLM Interaction}

\author{
 \textbf{Lvhua Wu\textsuperscript{1,2}\thanks{Equal contribution.}},
 \textbf{Xuefeng Jiang\textsuperscript{1,2}\footnotemark[1]},
 \textbf{Sheng Sun\textsuperscript{1}},
 \textbf{Yan Lei\textsuperscript{1,2}},
 \textbf{Tian Wen\textsuperscript{1,2}},
 \textbf{Yuwei Wang\textsuperscript{1}},
 \textbf{Min Liu\textsuperscript{1,2}\thanks{Corresponding author.}}
\\
\\
 \textsuperscript{1}Institute of Computing Technology, Chinese Academy of Sciences
\\
 \textsuperscript{2}University of Chinese Academy of Sciences
\\
 \texttt{\{wulvhua24s,jiangxuefeng21b,liumin\}@ict.ac.cn}
}

\begin{document}
\maketitle
\begin{abstract}
The rapid spread of fake news threatens social stability and public trust, highlighting the urgent need for its effective detection.
Although large language models (LLMs) show potential in fake news detection, they are limited by knowledge cutoff and easily generate factual hallucinations when handling time-sensitive news.
Furthermore, the thinking of a single LLM easily falls into early stance locking and confirmation bias, making it hard to handle both content reasoning and fact checking simultaneously.
To address these challenges, we propose ZoFia, a two-stage zero-shot fake news detection framework.
In the first retrieval stage, we propose novel Hierarchical Salience and Salience-Calibrated Minimum Marginal Relevance (SC-MMR) algorithm to extract core entities accurately, which drive dual-source retrieval to overcome knowledge and evidence gaps.
In the subsequent stage, a multi-agent system conducts multi-perspective reasoning and verification in parallel and achieves an explainable and robust result via adversarial debate.
Comprehensive experiments on two public datasets show that ZoFia outperforms existing zero-shot baselines and even most few-shot methods.
Our code has been open-sourced to facilitate the research community at \url{https://github.com/SakiRinn/ZoFia}.
\end{abstract}

\section{Introduction}

The rapid spread of fake news through social networks has become prevalent, posing severe threats to key domains such as politics \cite{fisher2016pizzagate}, economy \cite{bakir2018fake}, and livelihood \cite{zhou2020survey}.
The swift advancement of generative models further exacerbates this concern \cite{chen2023can,genfend,fairD,LiIE}.
In this context, developing effective, efficient and interpretable fake news detection methods is indispensable.

Early studies mainly rely on supervised learning \cite{monti2019fake,kaliyar2021fakebert} to train detection models, but their dependence on large-scale labeled data makes it hard to adapt to emerging news topics \cite{hoy2022exploring}.
Large language models (LLMs), with broad pretraining knowledge and strong contextual understanding \cite{su2023adapting}, significantly advance few-shot and zero-shot methods, providing new opportunities to overcome this bottleneck. In few-shot methods, LLMs either serve as auxiliary tools to perform data augmentation for downstream classifiers \cite{badactor}, or act directly as detectors through prompt learning \cite{jiang2022fake} and instruction tuning \cite{pavlyshenko2023analysis}, but they still do not fully escape reliance on training data. By contrast, zero-shot methods that do not require labeled samples are a more promising paradigm, which mainly guide LLM reasoning by context engineering \cite{zhang2023towards} and agentic architectures \cite{li2024large}.

However, zero-shot methods face reliability challenges. On the one hand, the internal knowledge of LLMs is limited by knowledge cutoff \cite{cheng2024dated}. Without the external information, they easily produce factual hallucinations \cite{ji2023survey} when handling dynamic news events. On the other hand, LLMs tend to lock their stance early \cite{echterhoff2024cognitive}, where confirmation bias \cite{wan2025unveiling} drives subsequent reasoning to merely rationalize the initial judgment \cite{wang2025truth}. Consequently, when external retrieval is introduced into a single LLM, it easily takes a cognitive shortcut that substitutes whether information can be retrieved for factual veracity, which seriously weakens content reasoning on the original news. This effect becomes much stronger when the retrieved evidence is irrelevant or contradictory \cite{tan2024blinded}. Therefore, we observe that the inherent reasoning defects of a single LLM prevent it from performing both news content reasoning and external evidence verification well simultaneously.

We argue that content reasoning and fact checking should be decoupled and reliable judgment can be achieved through multi-agent interaction. Based on this motivation, we propose ZoFia, a two-stage zero-shot framework for fake news detection. In the first entity-guided retrieval stage, we introduce Hierarchical Salience to address semantic dilution \cite{hou2021improving} of news entities, which fully leverages global and local semantics to score entity salience, and design a novel Salience-Calibrated Minimum Marginal Relevance (SC-MMR) algorithm to accurately extract core entities.
These entities then drive dual-source retrieval from Wikipedia and Open Web to mitigate knowledge cutoff and effectively suppress hallucinations. In the subsequent multi-LLM interaction stage, ZoFia assigns agents to perform content reasoning and fact checking in parallel and breaks stance locking of a single LLM via adversarial debate, ensuring the robustness and interpretability of the final verdict.

To sum up, our main contributions are outlined as follows:
\begin{itemize}[noitemsep, topsep=2pt, leftmargin=1em, rightmargin=1em]
    \item We propose ZoFia, a retrieval-augmented multi-agent zero-shot framework for fake news detection that effectively overcomes the inherent reasoning flaws of a single LLM.
    \item We introduce a novel granularity-aware metric Hierarchical Salience and design SC-MMR algorithm to extract news entities accurately for efficient retrieval.
    \item Comprehensive experiments on two public datasets demonstrate that ZoFia outperforms existing zero-shot baselines and even most few-shot methods.
    \item We open-source our code to facilitate the research community at \url{https://github.com/SakiRinn/ZoFia}.
\end{itemize}

\begin{figure*}[htbp]
    \centering
    \includegraphics[width=1.0\linewidth]{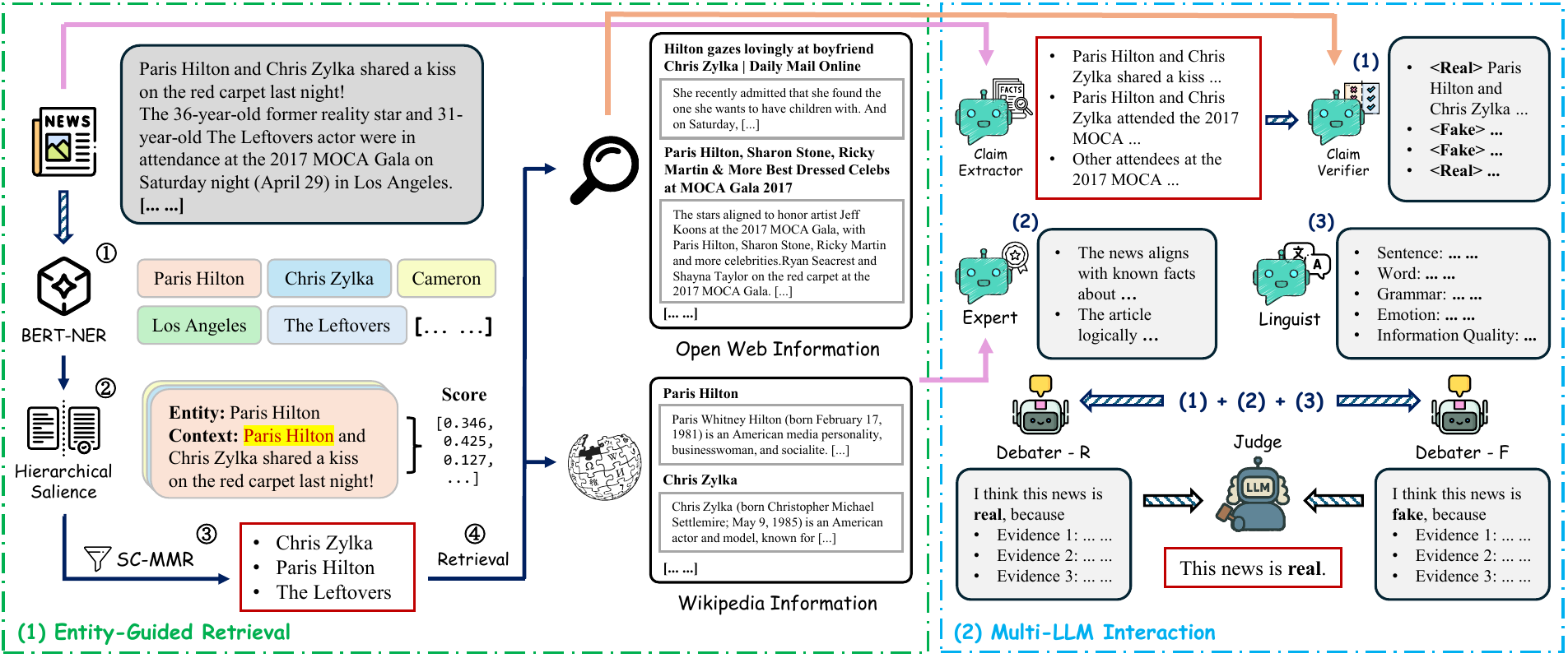}
    \caption{Overall architecture of our proposed ZoFia framework.}
    \label{fig:zofia}
\end{figure*}

\section{Related Work}

\textbf{LLM in Fake News Detection.}
The application of Large Language Models (LLMs) has become a research frontier in fake news detection, primarily divided into few-shot and zero-shot paradigms. In few-shot methods, researchers often use LLMs as auxiliary tools for data augmentation \cite{badactor, genfend}, or as detectors via post-training \cite{jiang2022fake,pavlyshenko2023analysis}. However, these methods fail to completely eliminate dependence on labeled data.
Zero-shot methods directly guide model reasoning through context engineering \cite{zhang2023towards} and agentic architectures \cite{li2024large, liu2024detect}, making judgments without labeled samples. Nevertheless, these methods fail to address the confirmation bias arising from a single reasoning chain. We decouple retrieval from reasoning via a multi-role multi-agent system and finally aggregate all evidence for judgment, achieving more comprehensive and robust zero-shot discrimination.

\textbf{Multi-Agent System.}
Multi-Agent Systems (MAS) have emerged as an effective paradigm to enhance LLMs for complex tasks. The pioneering work Chateval \cite{chan2023chateval} demonstrates that MAS improves both the robustness and accuracy of generation tasks.
Subsequent studies introduce this paradigm to reasoning tasks. For instance, COLA \cite{cola} designs a collaboration framework for stance detection, but it remains limited to analyzing the original text.
TruEDebate \cite{liu2025truth} applies structured debate to fake news detection, but its implementation tends to cause premature consensus convergence.
Recent agentic systems \cite{zhang2026expseek} also integrate reasoning more tightly with external knowledge access.
Our ZoFia introduces external information retrieval and independent modular analysis. This design aims to fundamentally mitigate knowledge cutoff and ensure the diversity and independence of arguments and analyses.

\section{Stage 1: Entity-Guided Retrieval}
\label{sec:retri}

This stage aims to acquire reliable external knowledge and instant factual evidence for the subsequent multi-agent system. It consists of four sequential modules. The first three modules precisely extract a set of core entities from the original news. These entities serve as keywords and are concatenated into a query for the final retrieval module, which retrieves from Open Web and Wikipedia.

\subsection{Entity Extractor}

This module first uses a pre-trained BERT-NER \cite{tjong-kim-sang-de-meulder-2003-introduction} model to perform named entity recognition (NER) on the news text. We adopt a lightweight BERT-NER model to obtain stable span-level confidence scores for dynamic filtering. This process can be expressed as:
\begin{equation}
\{(t_i, e_i, c_i)\}_{i=1}^N = \mathcal{M}_{\text{BERT-NER}}(T),
\end{equation}
where $\mathcal{M}$ is the pre-trained model, $T$ is the input news text.  $(t_i, e_i, c_i)$ denotes the recognized entity triplet, $t_i$ is the entity token, $e_i$ is the entity label, and $c_i$ is the confidence score for the corresponding label, expressed as the conditional probability $c_i=P(e_i | t_i, T; \mathcal{M}_\text{BERT-NER})$.

Due to the large number of recognized entities, we aggregate consecutive entity tokens to form new entity units $U_e$. Its confidence score $c(U_e)$ is calculated by averaging the confidence scores of all tokens in $U_e$, expressed as:
\begin{equation}
\small
c(U_e) = \frac{1}{|U_e|} \sum_{t_i \in U_e} c_i.
\end{equation}
We establish a dynamic confidence threshold to select entities with high confidence scores. Starting from an initial confidence score $\lambda_{\text{init}}$, if the number of selected entities fails to meet the predetermined minimum $n_{\text{min}}$, the algorithm iteratively lowers the confidence score by $\Delta\lambda$ and repeats the selection, until
at least $n_{\text{min}}$ entities are obtained.

\subsection{Salience Scorer}

This module aims to accurately quantify the importance of news entities, namely Entity Salience (ES) \cite{dunietz2014new}. Although the prior study \cite{bullough2024predicting} has demonstrated that bi-encoder architectures can efficiently estimate entity salience, their performance is often limited by semantic dilution \cite{hou2021improving} problem, which leads to severe underestimation of the importance of key entities.

We propose a novel \textbf{Hierarchical Salience} that avoids a single, coarse evaluation between an entity and the whole text. It decomposes an entity's overall importance into two orthogonal and multiplicative components: Local Salience $\mathcal{S}_{\text{local}}$ and Global Salience $\mathcal{S}_{\text{global}}$. Local Salience measures the semantic alignment between the entity and its immediate context, and Global Salience measures how much the local context contributes to the text's main content.

Formally, for a news text $ T $ that is an ordered sequence of sentences, consider any entity $ U_i $ that appears in sentence $ s_j $. We define its local context as $ \mathcal{C}(U_i) = s_{j-1} \oplus s_j \oplus s_{j+1} $. Aligned with \cite{bullough2024predicting}, we use a unified Sentence-BERT (SBERT) \cite{reimers2019sentence} encoder $ \mathcal{M}_{\text{SBERT}}(\cdot) $ to embed the entity $ U_i $, its local context $ \mathcal{C}(U_i) $, and the full text $ T $ to vectors $ \mathbf{v}_{U_i} $, $ \mathbf{v}_{\mathcal{C}(U_i)} $, and $ \mathbf{v}_{T} $. The hierarchical salience $\mathcal{S}_{\text{hier}}(U_i) $ is derived from the product of these two components as
\begin{equation}
\small
\mathcal{S}_{\text{hier}}(U_i) =
\underbrace{\frac{\mathbf{v}_{U_i} \cdot \mathbf{v}_{\mathcal{C}(U_i)}}{\|\mathbf{v}_{U_i}\| \cdot \|\mathbf{v}_{\mathcal{C}(U_i)}\|}}_{\mathcal{S}_{\text{local}}(U_i \mid \mathcal{C}(U_i))}
\cdot
\underbrace{\frac{\mathbf{v}_{\mathcal{C}(U_i)} \cdot \mathbf{v}_T}{\|\mathbf{v}_{\mathcal{C}(U_i)}\| \cdot \|\mathbf{v}_T\|}}_{\mathcal{S}_{\text{global}}(\mathcal{C}(U_i) \mid T)}.
\end{equation}
Hierarchical Salience provides a finer and more robust estimate of each entity's importance to the news content. It serves as the key criterion for entity filtering in subsequent modules.

\subsection{Keyword Selector}

This module aims to select an informative subset of keywords from the candidate entity set $\mathcal{U}_{\text{selected}}$ to mitigate the query drift \cite{carpineto2012survey} problem. However, this process faces two practical challenges. First, the hierarchical salience score $\mathcal{S}_{\text{hier}}(U_i)$ is highly sensitive to context granularity, so a fixed screening threshold is not effective. Second, coreference in news introduces high-scoring entities with repeated semantics, which harms the diversity of the keyword set.

To optimize relevance and diversity under these constraints, we propose an improved MMR \cite{carbonell1998use} algorithm, namely \textbf{Salience-Calibrated MMR (SC-MMR)}. At the $k$-th iteration, SC-MMR evaluates the score of a candidate entity $U_i$ by

\begin{subequations}
\small
\begin{gather}
\text{MMR}(U_i) =\lambda_k \mathcal{S}_\text{hier}(U_i)
- (1-\lambda_k) \max_{U_j \in \mathcal{U}_{\text{selected}}} \mathcal{S}(U_i, U_j),
\\
\mathcal{S}(U_i, U_j) = \frac{\mathbf{v}_{U_i} \cdot \mathbf{v}_{U_j}}{\|\mathbf{v}_{U_i}\| \,\|\mathbf{v}_{U_j}\|}.
\end{gather}
\end{subequations}
The key innovation of SC-MMR is to introduce a weight schedule $\lambda_k$ that changes with the number of selected keywords k, so that the focus gradually shifts from relevance to diversity. We adopt an annealing schedule with a lower bound:
\begin{equation}
\label{eq:lambda_k}
    \lambda_k = \max(\lambda_{\min}, \lambda_{\max} - \exp(\alpha\cdot k - \beta)).
\end{equation}
This form ensures that $\lambda_k$ decreases monotonically with $k$ while retaining a non-zero salience weight via $\lambda_{\min}$ to prevent the diversity term from fully dominating.

We further introduce a dynamic termination rule based on the relative change of the MMR score. The iteration continues only when the next best candidate $\mathcal{U}_{k+1}^*$ satisfies $\text{MMR}(\mathcal{U}_{k+1}^*) > \gamma \cdot \text{MMR}(\mathcal{U}_{k}^*)$, where $\gamma$ is a decay factor.
This design prevents the decline in entity quality caused by diminishing marginal utility.

\subsection{Information Retrieval}

This module uses the keyword set $\mathcal{K} = \{k_1, k_2, \ldots, k_N\}$ distilled in previous modules to build a comprehensive external knowledge base $\mathcal{E}$, serving as supplementary context for the subsequent detection stage. We implement a dual-source retrieval that gathers information from the Open Web and Wikipedia in parallel, ensuring both timeliness and authority.

\textbf{Retrieval from Open Web.}
We compose all keywords into an aggregated query for the Open Web $Q_{\text{web}}$ with the following logic:
\begin{equation}
\small
    Q_{\text{web}} = \left( \bigwedge_{i=1}^{N} k_i \right)
\end{equation}
The operator $\land$ requires that results relate to all keywords, which enables strict cross-keyword verification. We explicitly exclude news and Wikipedia sources. The former avoids retrieving duplicate reports to prevent source contamination \cite{deng2023investigating}, and the latter prevents redundancy with the subsequent Wikipedia retrieval. For each returned page, we only extract its summary and search snippet as the raw corpus.

\textbf{Retrieval from Wikipedia.}
The summary section of a Wikipedia entry usually provides the most precise definition for an entity. However, a single keyword $k_i$ often corresponds to multiple Wikipedia entries, which introduces semantic ambiguity. We build a \textbf{context-aware disambiguation} mechanism that uses the original local context $\mathcal{C}(U_i)$ of keyword $k_i$ in the news text to perform accurate matching.

When Wikipedia returns a list with $M$ candidate senses $\mathcal{O}(k_i) = \{o_1, o_2, \ldots, o_M\}$, the mechanism examines each candidate $o_m$. It constructs a temporary modified context $\mathcal{C}(U_i \leftarrow o_m)$ by replacing the original entity $U_i$ with the description text of $o_m$. The optimal sense $o_i^*$ of $U_i$ is defined as the option that maximizes the cosine similarity between the vector of the original context and the vector of the modified context:
\begin{equation}
\small
o_i^* = \mathop{\arg\max}_{o_m \in \mathcal{O}(k_i)}
\frac{\mathbf{v}_{\mathcal{C}(U_i)} \cdot \mathbf{v}_{\mathcal{C}(U_i \leftarrow o_m)}}{\|\mathbf{v}_{\mathcal{C}(U_i)}\| \|\mathbf{v}_{\mathcal{C}(U_i \leftarrow o_m)}\|}.
\end{equation}
where $\mathbf{v}$ denotes the context vectors embedded by the pretrained SBERT $\mathcal{M}_{\text{SBERT}}$. After identifying the unique entry, we extract the first 3 sentences of its summary as supplementary material.

\section{Stage 2: Multi-LLM Interaction}

The external information provided by entity-guided retrieval and the prior knowledge of LLMs builds a \textbf{Multi-Source Information Matrix}.
This stage employs a dual-state multi-LLM system that fully exploits this matrix to perform parallel and multi-perspective content reasoning and claim verification, which are finally aggregated to reach a robust and interpretable judgment.

This stage operates in two orthogonal interaction states. \textbf{LLM Collaboration} performs parallel analyses across multiple agents to reduce inferential variance. \textbf{LLM Debate and Judgment} introduces an adversarial debate to reduce systemic bias. These two states form a complete reasoning chain from mining divergent evidence to making a convergent judgment.

\subsection{LLM Collaboration}

\begin{figure}[htbp]
    \centering
    \includegraphics[width=0.9\linewidth]{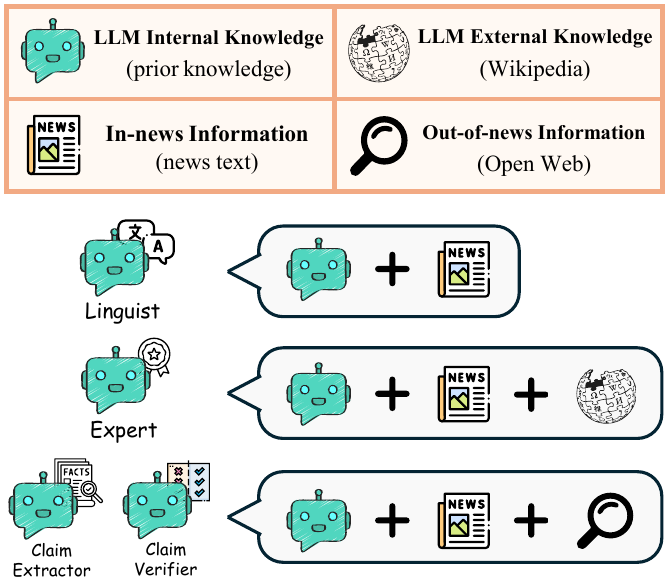}
    \caption{The diagram of Multi-Source Information Matrix and the quadrants used by LLM agents.}
    \label{fig:matrix}
\end{figure}

LLM Collaboration state aims to reduce inferential variance. Through parallel analysis by multiple agents, it transforms the \textbf{Multi-Source Information Matrix} from the previous stage into a structured evidence pool, which provides a stable decision basis for the subsequent adversarial debate. As illustrated in Figure \ref{fig:matrix}, the matrix systematically integrates 4 information sources that are orthogonal and complementary in evidence distribution:

\begin{itemize}[noitemsep, topsep=2pt, leftmargin=1em, rightmargin=1em]
    \item \textit{In-news Information:} Comes from the original text of the target news and provides the basic core content.
    \item \textit{Out-of-news Information:} Comes from Open Web retrieval and provides the most timely and broadest materials.
    \item \textit{LLM internal Knowledge:} Comes from the model's prior knowledge and provides generalized common sense.
    \item \textit{LLM external Knowledge:} Comes from Wikipedia retrieval and provides the most precise summary for the core entities.
\end{itemize}

The subsequent analysis decomposes the tasks and assigns them to agents with distinct roles, so that each agent focuses on a specific quadrant of the matrix for specialized processing.

\subsubsection{Linguist}
Following prior studies \cite{shahid2022detecting}\cite{zhou2018fake}, the linguist agent is designed to systematically divide the text into 5 linguistic dimensions that are strongly associated with misinformation:

\begin{itemize}[noitemsep, topsep=2pt, leftmargin=1em, rightmargin=1em]
    \item \textit{Sentence:} Lexical complexity, sentence length, and formality of tone.
    \item \textit{Word:} Frequency of superlatives, affective language, and pronoun distribution.
    \item \textit{Grammar:} Patterns of reported speech, passive voice, and negation.
    \item \textit{Emotion:} Affective terms in the text and the headline, and the degree of incendiary tone.
    \item \textit{Information Quality:} Presence of clickbait, information overload, or context mismatch.
\end{itemize}

To maintain objective independence, each dimension is evaluated in an isolated session. For each dimension, the LLM explicitly indicates whether it reflects the news is real or fake. It uses 2 quadrants of the Multi-Source Information Matrix: LLM internal knowledge and in-news information.

\subsubsection{Domain-Specific Expert}
It operates in a dynamic and adaptive manner. The system first identifies the most relevant domain from the news text and assigns the agent a precise expert role, such as "economist" or "journalist".

The expert with this role then analyzes along the following 2 dimensions:

\begin{itemize}[noitemsep, topsep=2pt, leftmargin=1em, rightmargin=1em]
    \item \textit{Knowledge Concordance}: Examine all claims, viewpoints, and details for sound reasoning; identify departures from common sense.
    \item \textit{Logical Integrity}: Examine argument-to-conclusion coherence with domain-specific common sense; identify logical errors or unsupported leaps.
\end{itemize}

It uses 3 quadrants of the Multi-Source Information Matrix: in-news information, LLM internal knowledge, and LLM external knowledge.

\subsubsection{Claim Verification}

Existing research \cite{niu2024veract} demonstrates that claim-based fact checking effectively serves as a reference for LLM-based detection. In our system, a serial pipeline composed of \textbf{Claim Extractor} and \textbf{Claim Verifier} deconstructs and verifies factual claims in the news text. These 2 agents use 3 quadrants of the Multi-Source Information Matrix: LLM internal knowledge, in-news information, and out-of-news information from Open Web.

\textbf{Claim Extractor.}
The $\mathcal{M}_\text{extractor}$ agent converts unstructured news text $T$ into a set of verifiable structured claims. Its function is formalized as follows:

\begin{equation}
    \mathcal{M}_\text{extractor}(T) \rightarrow \{ q_{\text{core}}, \{q_{\text{sub}_1}, \dots, q_{\text{sub}_m}\} \}
\end{equation}

where $q_{\text{core}}$ is the core claim that determines the veracity of the news, and ${q_{\text{sub}}}$ is a collection of supporting subclaims. All outputs are restated as concise and objective declarative sentences.

\textbf{Claim Verifier.}
This agent verifies each claim $q$ independently. To ensure precision and control costs \cite{purwar2024evaluating}, a simple retrieval-augmented generation (RAG) \cite{lewis2020retrieval} is implemented to build a highly relevant context $\mathcal{C}_{\text{rel}}(q)$ for each $q$ from the Open Web corpus $\mathcal{I}_{\text{web}}$.

The context consists of text chunks $c_j$ whose cosine similarity with the claim representation $\mathbf{v}_q$ exceeds a threshold $\theta_{\text{sim}}$:

{\small
\begin{equation}
\mathcal{C}_{\text{rel}}(q) = \{ c_j \in \text{top-k}(\mathcal{I}_{\text{web}}, q) \mid \frac{\mathbf{v}_q \cdot \mathbf{v}_{c_j}}{\|\mathbf{v}_q\| \|\mathbf{v}_{c_j}\|} \ge \theta_{\text{sim}} \}.
\end{equation}
}

All decisions are strictly based on $\mathcal{C}_{\text{rel}}(q)$. The final output includes a clear label (\textit{"Supports"}, \textit{"Refutes"}, or \textit{"Not Enough Information"}), a brief reasoning, and evidence directly quoted from $\mathcal{C}_{\text{rel}}(q)$ to suppress hallucinations.

\subsection{LLM Debate and Judgment}

LLM Debate and Judgment state aims to suppress systemic bias. We introduce a multi-round adversarial debate framework that forces LLMs to explore both supporting and refuting views of the truthfulness of news equally. This design directly mitigates the thought degeneration (DoT) \cite{liang2023encouraging} phenomenon that often appears in a single linear reasoning chain.

\begin{algorithm}[htbp]
\caption{Dynamic Adversarial Debate}
\label{alg:debate}
\KwIn{
$\mathbb{E}$: The complete evidence pool
}
\KwOut{$D$: The final decision $\{\textit{Real}, \textit{Fake}\}$}

$\mathbb{H} \gets \emptyset$; $A_{\text{con}} \gets \text{null}$; $D \gets \textit{Insufficient}$\;

\While{$D = \text{Insufficient}$}{
    $A_{\text{pro}} \gets \mathcal{M}_{\text{pro}}.\text{GenerateArgument}(\mathbb{E}, A_{\text{con}})$\;
    $A_{\text{con}} \gets \mathcal{M}_{\text{con}}.\text{GenerateArgument}(\mathbb{E}, A_{\text{pro}})$\;
    $\mathbb{H} \gets \mathbb{H} \cup \{(A_{\text{pro}}, A_{\text{con}})\}$\;
    $D \gets \text{Judge.Assess}(\mathbb{H})$\;
}
\Return $D$\;
\end{algorithm}

As shown in Algorithm \ref{alg:debate}, a pair of debate agents $\mathcal{M}_{\text{pro}}$ and $\mathcal{M}_{\text{con}}$ act as opposing reasoners based on the evidence pool $\mathbb{E}$. In each round, the active debater first rebuts the opponent's last argument and then presents a new argument. After each exchange, a judge agent evaluates the debate history $\mathcal{H}$ and outputs a ternary judgment $D$.

This dynamic termination mechanism ensures that the debate stops once the information is sufficient for decision, effectively balancing the depth and efficiency of reasoning. The debate history $\mathbb{H}$ provides a traceable reasoning chain composed of pro. and con. arguments, making the final judgment highly interpretable.

\begin{table*}[htbp]
\centering
\caption{Accuracy (Acc.) and F1-Score (F1) comparison of few-shot / zero-shot methods on PolitiFact and GossipCop. The \textbf{bold} and \underline{underlined} denote the best and second-best performance.}
\label{tab:acc-f1}
\begin{adjustbox}{max width=\linewidth}
\footnotesize
\begin{tabular}{c|c|c|cc|cc}
\toprule \toprule
\multirow{2}{*}{\makecell[c]{\textbf{Category}}} & \multirow{2}{*}{\textbf{Method}} & \multirow{2}{*}{\textbf{LLM Usage}} & \multicolumn{2}{c|}{\textbf{PolitiFact}} & \multicolumn{2}{c}{\textbf{GossipCop}} \\

\cmidrule(lr){4-5} \cmidrule(l){6-7}
 & & & \textbf{Accuracy} & \textbf{F1-Score} & \textbf{Accuracy} & \textbf{F1-Score} \\

\midrule
\multirow{4}{*}{Few-shot}
& PSM \cite{nan2021mdfend} & Non-LLM & 70.00 & 49.15 & 77.44 & 41.73 \\
& MDFEND \cite{nan2021mdfend} & Non-LLM & 65.50 & 62.30 & 41.27 & 40.20 \\
& KPL \cite{jiang2022fake} & Non-LLM & 58.33 & 60.40 & 42.71 & 42.08 \\
\cmidrule(lr){2-7}
& ARG \cite{hu2024bad} & LLM-assisted & 74.00 & 67.16 & 61.41 & 42.32 \\
& DKFND \cite{liu2024detect} & LLM-assisted & 87.00 & 82.43 & \textbf{82.37} & 55.22 \\

\midrule
\multirow{9}{*}{\makecell[c]{Zero-shot}}
& Auto-CoT \cite{zhang2022automaticchainthoughtprompting} & LLM-based & \underline{89.65} & 73.67 & 60.01 & 48.15 \\
& ReAct (Search API) \cite{yao2023reactsynergizingreasoningacting}    & LLM-based & 74.73 & 67.64 & 74.03 & 47.30 \\
& HiSS \cite{zhang2023towards} & LLM-based & 64.82 & 56.80   & 68.81 & 40.40 \\
& Web Retrieval Agents \cite{tian2024web} & LLM-based & 77.83 & 64.88   & 66.62 & 46.54 \\
& FactAgent \cite{li2024large} & LLM-based & 80.59 & 70.06   & 73.80 & 56.22 \\
& CAPE-FND \cite{jin2023capefnd} & LLM-based & 76.81 & 62.43 & 63.79 & 49.17 \\

\cmidrule(lr){2-7}
& DeepSeek-v3 \cite{deepseekai2024deepseekv3technicalreport} & Only-LLM & 78.99 & 40.24 & 61.03 & 28.17 \\
& GPT-4o-mini \cite{openai_gpt4omini} & Only-LLM & 73.58 & 41.66 & 66.73 & 33.18 \\
& Qwen3-32B \cite{qwen3technicalreport} & Only-LLM & 76.57 & 43.35 & 63.82 & 32.45 \\

\cmidrule(lr){2-7}
& \textbf{ZoFia (DeepSeek-v3)} & LLM-based  & \textbf{91.52} & \textbf{87.88} & \underline{79.04} & \textbf{61.22} \\
& \textbf{ZoFia (GPT-4o-mini)} & LLM-based  & 75.28 & 75.19 & 68.90 & 56.20 \\
& \textbf{ZoFia (Qwen3-32B)} & LLM-based  & 84.06 & \underline{81.76} & 69.71 & \underline{59.07} \\

\bottomrule
\end{tabular}
\end{adjustbox}
\end{table*}

\section{Experiment}
\subsection{Experimental Setting}


\textbf{Datasets.}
Following previous state-of-the-art work \cite{liu2024detect}, our experiments are conducted on two widely recognized fake news datasets: GossipCop and PolitiFact \cite{shu2020fakenewsnet}. GossipCop focuses on entertainment, mainly Hollywood celebrity news; PolitiFact focuses on politics, drawing on fact-checks of U.S. political figures.
Considering that some links in the initial dataset have become invalid, we adopt the available version publicly re-released in \cite{su2023adapting}.

\textbf{Metrics.}
We use accuracy and macro F1-Score as evaluation metrics. F1-Score is less affected by data imbalance, so it serves as the primary metric for assessment.

\textbf{Baselines.}
We incorporate two groups of baselines. The first group includes advanced few-shot methods: PSM \cite{ni2020improving}, MDFEND \cite{nan2021mdfend}, ARG \cite{hu2024bad}, DKFND \cite{liu2024detect}, and KPL \cite{jiang2022fake}, with reliable metrics from the existing works \cite{jin2024mm,hu2024multi,liu2024detect}. The second group consists of representative zero-shot methods for comparison: Auto-CoT \cite{zhang2022automaticchainthoughtprompting}, ReAct \cite{yao2023reactsynergizingreasoningacting} equipped with a search API, HiSS \cite{zhang2023towards}, Web Retrieval Agents \cite{tian2024web}, FactAgent \cite{li2024large}, and CAPE-FND \cite{jin2023capefnd}.

We implement ZoFia based on {\small\texttt{DeepSeek-v3}} \cite{deepseekai2024deepseekv3technicalreport}, {\small\texttt{GPT-4o-mini}} \cite{openai_gpt4omini} and Qwen3-32B \cite{qwen3technicalreport}, and compare it with only-LLM inference. All other LLM-based zero-shot methods are based on {\small\texttt{DeepSeek-v3}}.

\textbf{Implementation details.} We set the dynamic threshold of entity extraction as $\lambda_{\text{init}}=0.8$ and $\Delta\lambda=0.1$ and the decay factor of MMR as $\gamma=0.5$. The maximum number of entries for Open Web retrieval is 10; for Wikipedia, we retrieve the first 3 sentences for each entry. The minimum similarity threshold for claim extraction is $\theta_{\text{sim}}=0.1$. The NER model is selected as {\small\texttt{dslim/bert-base-NER}} \cite{tjong-kim-sang-de-meulder-2003-introduction}, and the SBERT model is selected as {\small\texttt{all-MiniLM-L6-v2}} \cite{reimers2019sentence}. For all base LLMs, we set the temperature to 0.

Brave API\footnote{\url{https://brave.com/search/api/}} is selected as Open Web retrieval API. To strictly prevent label information leakage, we set the search cutoff date to the day before the publication date of each sample URL and exclude dataset-associated fact-checking domains from open-web retrieval (gossipcop.com, politifact.com, and snopes.com).

\subsection{Main Experiment}

ZoFia demonstrates exceptional performance on both datasets as shown in Table \ref{tab:acc-f1}, consistently outperforming existing zero-shot and few-shot baselines. On the fact-intensive PolitiFact dataset, ZoFia's advantages are particularly pronounced, with its accuracy (91.52\%) and F1-Score (87.88\%) not only far exceeding other zero-shot methods but also surpassing all few-shot baselines by a substantial performance margin.

The detection task on the GossipCop dataset presents greater challenges, as its content often proves difficult to verify due to subjectivity and factual ambiguity. In this scenario, ZoFia achieves the highest F1-Score (61.22\%), surpassing all baseline models. Although DKFND attains slightly higher accuracy, its F1-Score is notably lower. In contrast, ZoFia's superior F1-Score demonstrates more balanced and robust detection performance across both true and fake news categories while maintaining high precision.

The results clearly show that as a zero-shot method, ZoFia consistently outperforms all zero-shot baselines and even most few-shot methods. Compared to only-LLM inference, ZoFia provides stable and substantial performance gains, demonstrating the superiority of its architecture.

\subsection{Ablation Study}
To assess the contribution of each component in ZoFia, we conduct ablation studies on GossipCop with DeepSeek-v3 and Qwen3-32B. As shown in Table \ref{tab:ablation}, removing Open Web retrieval causes clear degradation in both settings, which shows that timely external evidence is essential for both base models. Removing all retrieval sources further lowers F1-Score to 58.26\% on DeepSeek-v3 and 50.85\% on Qwen3-32B. In LLM collaboration, removing either the linguist or the expert weakens the framework, and the expert contributes more; removing both leads to a larger cumulative drop. Removing debate also consistently reduces F1-Score, which shows that it helps mitigate single-perspective bias. The F1-Score drops by 2.06\% on DeepSeek-v3 and by 5.00\% on Qwen3-32B, and this larger drop suggests that the smaller model is more sensitive to the loss of this mechanism.

\begin{table}[t]
\centering
\caption{Ablation study results on GossipCop. (ACC.: Accuracy, F1: F1-Score)}
\label{tab:ablation}
\small
\begin{adjustbox}{max width=\linewidth}
\begin{tabular}{l|cc|cc}
\toprule
\toprule
\multirow{2}{*}{\textbf{Component}} & \multicolumn{2}{c|}{\textbf{DeepSeek-v3}} & \multicolumn{2}{c}{\textbf{Qwen3-32B}} \\
\cmidrule(lr){2-3} \cmidrule(lr){4-5}
& Acc. & F1 & Acc. & F1 \\
\midrule
ZoFia & 79.04 & 61.22 & 69.71 & 59.07 \\
\midrule
\textit{w/o} Wikipedia retrieval & 77.67 & 59.90 & 68.09 & 56.58 \\
\textit{w/o} Open Web retrieval & 71.71 & 58.14 & 64.18 & 52.47 \\
\textit{w/o} All retrievals & 71.37 & 58.26 & 62.65 & 50.85 \\
\midrule
\textit{w/o} Linguist analysis & 77.86 & 58.87 & 67.55 & 54.62 \\
\textit{w/o} Expert analysis & 76.77 & 57.32 & 66.87 & 54.39 \\
\textit{w/o} All analyses & 73.51 & 57.90 & 63.82 & 52.71 \\
\midrule
\textit{w/o} Claim verification & 79.23 & 60.50 & 67.48 & 55.17 \\
\textit{w/o} Debate & 75.01 & 59.16 & 67.13 & 54.07 \\
\bottomrule
\end{tabular}
\end{adjustbox}
\end{table}

\subsection{Effectiveness of Entity-Guided Retrieval}
To verify the effectiveness of entity-guided retrieval in ZoFia, we fix the second stage and vary only how the query is constructed on GossipCop with Qwen3-32B. TF-IDF \cite{sparck1972statistical}, PromptNER \cite{ashok2023promptner}, and Only-LLM NER still build queries from keywords, but replace the entity acquisition mechanism, and their F1-Scores drop by 5.69\%, 3.94\%, and 6.33\% relative to ZoFia. Sentence Salience \cite{bullough2024predicting}, Core Claim, and OR Composition instead use sentence-level queries or relax conjunctive matching from AND ($\wedge$) to OR ($\vee$), and their F1-Scores fall by 12.92\%, 10.26\%, and 6.81\%. These results show that the gain does not come from arbitrary keyword lists, and it depends on accurate entity extraction and strict entity-level constraints, which make the original conjunctive entity-guided query the most effective design.

\begin{table}[t]
\centering
\setlength{\tabcolsep}{6pt}
\caption{Effectiveness of query construction variants in ZoFia on GossipCop.}
\label{tab:query_eff}
\resizebox{\columnwidth}{!}{%
\begin{tabular}{lcc}
\toprule
\toprule
\makecell[c]{\textbf{Method}} & \textbf{Accuracy} & \textbf{F1-Score} \\
\midrule
ZoFia & 69.71 & 59.07 \\
ZoFia (\textit{w/o} All retrievals) & 62.65 & 50.85 \\
\midrule
ZoFia (TF-IDF) & 66.97 & 53.38 \\
ZoFia (PromptNER) & 63.08 & 55.13 \\
ZoFia (Only-LLM NER) & 61.69 & 52.74 \\
\midrule
ZoFia (Sentence Salience) & 68.64 & 46.15 \\
ZoFia (Core Claim) & 66.65 & 48.81 \\
ZoFia (OR Composition) & 64.89 & 52.26 \\
\bottomrule
\end{tabular}%
}
\end{table}

\subsection{Sensitivity Analysis of SC-MMR}
\begin{figure}[htbp]
    \centering
    \includegraphics[width=0.95\linewidth]{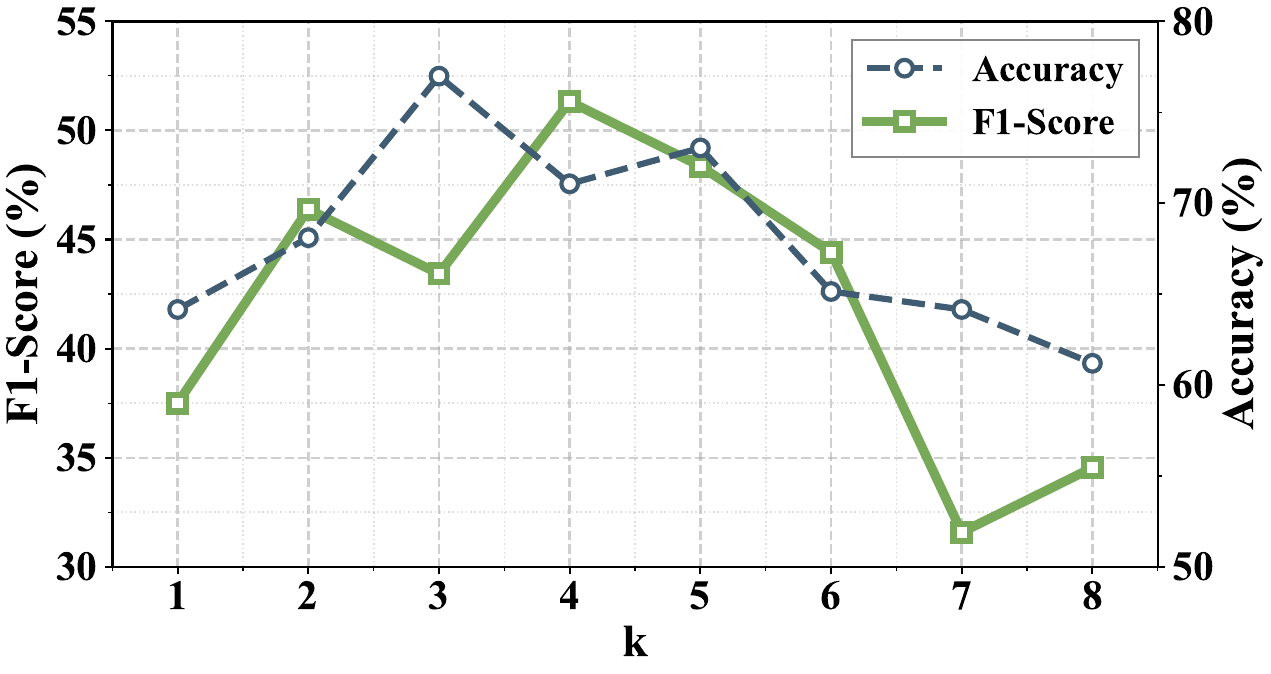}
    \caption{The effect of the number of keywords $k$ on performance (F1-Score).
}
    \label{fig:k}
\end{figure}

To motivate the design of the dynamic weight $\lambda_k$, we first conduct an experiment with the weight fixed at $\lambda=0.5$ to observe the direct impact of the keyword count $k$ on performance. As shown in Figure \ref{fig:k}, the F1-Score remains stable when $k \le 6$ but drops sharply beyond this point.

This phenomenon reveals an inflection point: when the number of keywords exceeds $6$, the risk of introducing noise and redundancy begins to outweigh the benefits of new information. It suggests that the strategic focus must shift from relevance to diversity near the critical inflection point of $k \approx 6$. Consequently, we adopt the annealing schedule form defined in Equation \ref{eq:lambda_k} and determine its parameters as $\alpha=0.3$ and $\beta=2.5$ to fit this downward trend.

\subsection{Efficiency of Token Utilization}
To investigate ZoFia's reasoning efficiency, a controlled comparison experiment is conducted. We provide CoT \cite{wei2022chain}, FactAgent \cite{li2024large}, and ZoFia with the same materials and impose a unified limit on output tokens to evaluate their performance. Since ZoFia cannot complete full reasoning below 400 tokens, we set its evaluation range to 400 tokens and above.

\begin{figure}[htbp]
    \centering
    \includegraphics[width=0.95\linewidth]{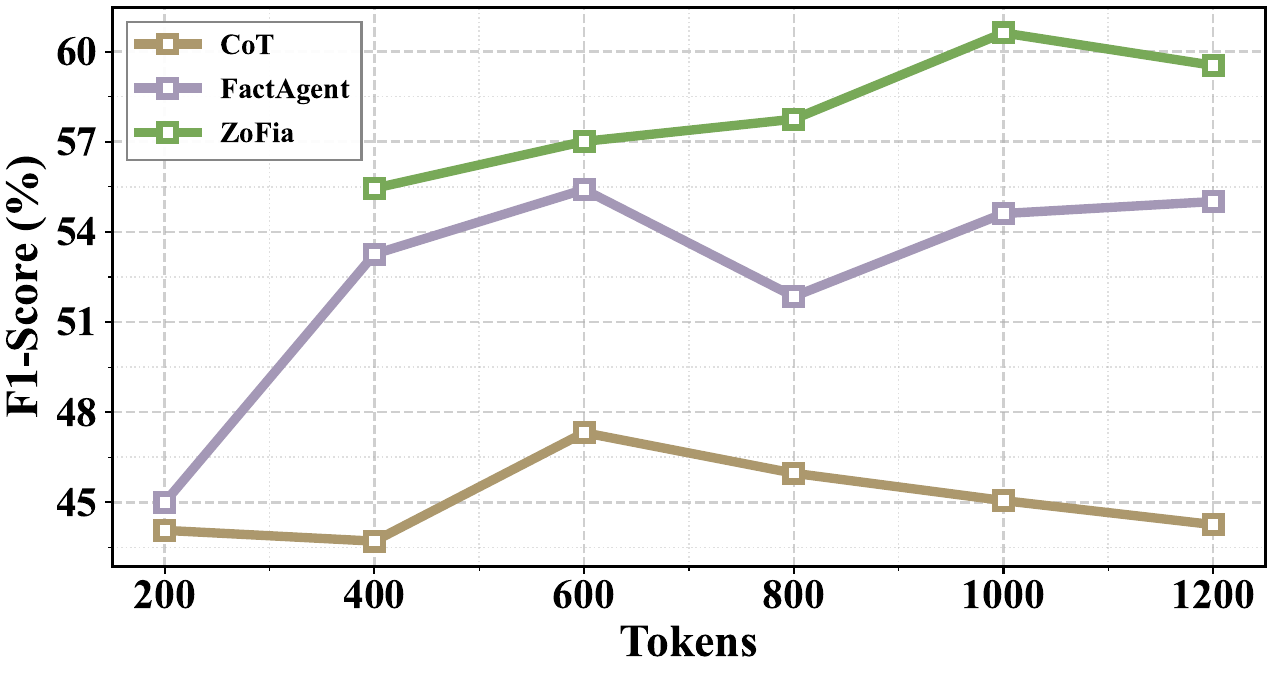}
    \caption{Performance (F1-Score) comparison of methods under maximum output token limits.}
    \label{fig:efficiency}
\end{figure}

The results are shown in Figure \ref{fig:efficiency}. At all budget points above 400 tokens, ZoFia achieves a clearly higher F1-Score than other baseline methods, which demonstrates high reasoning efficiency. In contrast to CoT and FactAgent, whose performance saturates after around 600 tokens, ZoFia shows a stable upward trend even at 1200 tokens. This indicates that ZoFia can consistently convert tokens into performance gains within an acceptable budget.

Note that this experiment only constrains output tokens, because prompt caching \cite{gim2024prompt} can extensively amortize input overhead, making output cost more critical for efficiency.

\section{Conclusion}
In this paper, we propose ZoFia, a retrieval-augmented multi-agent zero-shot framework for fake news detection, to address the cognitive conflict of a single LLM between content reasoning and fact checking. It first utilizes our novel Hierarchical Salience and Salience-Calibrated Minimum Marginal Relevance (SC-MMR) algorithm to accurately extract core entities from news text, which then guide dual-source retrieval from Open Web and Wikipedia. Next, the multi-agent system conducts analysis and verification in parallel and makes a final verdict via adversarial debate.
This process effectively reduces confirmation bias from a single reasoning perspective and ensures robust and explainable results. Comprehensive experiments on two public datasets show that ZoFia outperforms existing zero-shot baselines in both performance and efficiency.

\section*{Limitations}

Though ZoFia shows strong detection capabilities, its application and evaluation face multiple constraints. Due to copyright and privacy concerns, there has been a recent lack of high-quality, continuously updated public datasets.
It prevents us from evaluating ZoFia on the most recent news. Building the next generation of benchmark datasets that meet ethical standards and reflect real-world information dynamics is a critical step in this field.

The efficiency of using external retrieval can also be improved.
Since it is not our main focus, we integrate only a lightweight retrieval-augmented generation (RAG) module for claim verification.
Future work that adopts more advanced RAG architectures, such as a re-ranking model and more advanced mechanisms, may further strengthen the exploitation of external knowledge.
We plan to conduct more comprehensive benchmark evaluations of the detection capabilities of LLMs and multi-agent systems for fake news \cite{benchFake,kuntur2024fake,llm4cvd}.

ZoFia currently focuses on the text modality. Modern misinformation increasingly appears in multi-modal form that combines images and text.
Extending ZoFia to the multimodal domain has strong potential.
One direction is to introduce vision language models (VLMs) as a dedicated visual expert. Another is to study how different modalities interact during the debate process to achieve effective fusion.

\section*{Acknowledgments}
This work was supported by the National Natural Science Foundation of China (No. 62472410).
We thank the reviewers and chairs for their constructive feedback.



\bibliography{main}

\appendix

\section{Implementation Details of Modules in Entity-Guided Retrieval}

\subsection{Entity Extraction}

The goal of entity extraction is to provide a stable set of semantic anchors for subsequent retrieval. To prevent retrieval link failures caused by entity sparsity or confidence distribution shifts, we introduce dynamic confidence threshold filtering after extraction, maintaining a controllable balance between the quality and usability of the entity set.

Algorithm \ref{alg:dynamic-entity-filter} details the implementation of this process. The inputs are the news text $\mathcal{T}$, the Named Entity Recognition (NER) model $\mathcal{M}_{\text{NER}}$, an initial threshold $\tau_0$, and a threshold decay step $\Delta$. The output is the filtered entity set $\mathcal{E}_{\text{sel}}$. The algorithm first runs NER on the full text to obtain the raw entity set $\mathcal{E}_{\text{raw}}=\mathcal{M}_{\text{NER}}(\mathcal{T})$, where each entity $e$ comes with a model confidence score $e.\text{ner\_score}$ that characterizes the reliability of the entity boundary and type prediction. If $\mathcal{E}_{\text{raw}}$ is empty, the algorithm directly returns an empty set. This indicates that the text lacks identifiable entity signals or the model cannot provide reliable predictions, in which case the retrieval stage should degrade to coarser-grained information clues or skip the entity-guided strategy.

When $\mathcal{E}_{\text{raw}}$ is not empty, the algorithm employs dynamic threshold filtering to construct $\mathcal{E}_{\text{sel}}$. Specifically, it starts with a high initial threshold $\tau_0$ and retains only entities satisfying $e.\text{ner\_score}\ge \tau$, prioritizing entity quality and precision. If $\mathcal{E}_{\text{sel}}$ remains empty under the current threshold, the threshold decreases stepwise by $\Delta$ until the system selects at least one entity or the threshold drops to a lower bound of $0.1$. This mechanism reflects a clear constraint: entity extraction must provide a non-empty semantic entry point for subsequent retrieval; otherwise, retrieval degenerates into anchor-less generalized queries. Meanwhile, the lower bound threshold $0.1$ prevents the uncontrolled introduction of low-confidence noisy entities, thereby maintaining a baseline quality while ensuring usability.

Dynamic threshold filtering transforms entity selection from a fixed hyperparameter setting into a sample-adaptive process. This enables the system to cover both entity-dense and entity-sparse news narratives, providing a stable candidate foundation for subsequent Hierarchical Salience and SC-MMR.

\begin{algorithm}[htbp]
\caption{Entity Extraction and Dynamic-Threshold Filtering}
\label{alg:dynamic-entity-filter}
\KwIn{
$\mathcal{T}$: News text; \\
$\mathcal{M}_{\text{NER}}$: NER model; \\
$\tau_0$: Initial confidence threshold; \\
$\Delta$: Threshold decay step
}
\KwOut{$\mathcal{E}_{\text{sel}}$: Selected entity set}

$\mathcal{E}_{\text{raw}} \gets \mathcal{M}_{\text{NER}}(\mathcal{T})$\;
\If{$\mathcal{E}_{\text{raw}} = \emptyset$}{
    \Return $\emptyset$\;
}

$\tau \gets \tau_0$; $\mathcal{E}_{\text{sel}} \gets \emptyset$\;
\While{$\mathcal{E}_{\text{sel}}=\emptyset \ \wedge\ \tau \ge 0.1$}{
    $\mathcal{E}_{\text{sel}} \gets \{e \in \mathcal{E}_{\text{raw}} \mid e.\text{ner\_score} \ge \tau\}$\;
    $\tau \gets \tau - \Delta$\;
}
\Return $\mathcal{E}_{\text{sel}}$\;
\end{algorithm}

\subsection{SC-MMR Algorithm}

A larger set of query keywords is not necessarily better, as it defines the boundary of the external evidence space. SC-MMR is designed to select a set of core entities that are sufficiently informative yet minimally redundant within a limited scale, which makes subsequent retrieval more stable and controllable.

Algorithm \ref{alg:sc-mmr} presents the complete procedure of SC-MMR. The input is a mapping from entities to Hierarchical Salience scores $D:\{U_i \mapsto S_{\text{Hier}}(U_i)\}$, and a model $M_{\text{SBERT}}$ for calculating entity vector representations. The output is the final selected entity set $U_{\text{selected}}$. The algorithm first performs one-time pre-processing on the candidate entity set $U=\text{keys}(D)$. It uses $M_{\text{SBERT}}$ to encode each entity into a vector and constructs a pairwise similarity matrix $M_{\text{sim}}$, where $M_{\text{sim}}[i,j]$ measures the semantic proximity between entities $U_i$ and $U_j$. Pre-computing this matrix moves repetitive similarity calculations out of the iterative loop, so subsequent rounds only require table look-ups and maximum value operations.

The algorithm then enters the selection process. It first selects the entity with the highest salience score $D[U_i]$ from the candidate set $U_{\text{candidate}}$ as the seed $U^*$ and adds it to $U_{\text{selected}}$. This initialization ensures the first step is entirely driven by relevance. It then enters a `while` loop, where the goal of each round is to find the entity among the remaining candidates that yields the highest combined score of relevance and diversity. For any candidate $U_i$, the algorithm first calculates:
$$
\text{sim}_{\max}=\max_{U_j\in U_{\text{selected}}} M_{\text{sim}}[i,j],
$$
which represents the similarity between the candidate and the most similar entity in the current selected set, indicating the maximum redundancy that adding $U_i$ might introduce. Subsequently, it scores the candidate using:
$$
\text{MMR}_{\text{curr}}=\lambda_k\cdot D[U_i]-(1-\lambda_k)\cdot \text{sim}_{\max}
$$
Here, $\lambda_k$ is a weight schedule that varies with the number of selected entities $k=|U_{\text{selected}}|$, with a lower bound of 0.1. Intuitively, a larger $\lambda_k$ at small $k$ favors high-salience core entities to ensure factual coverage. As $k$ increases, $\lambda_k$ gradually decreases to strengthen the diversity penalty. This suppresses the repetitive inclusion of coreferential or semantically close entities, thereby mitigating query drift caused by keyword inflation. Each round iterates through $U_{\text{candidate}}$ to obtain the global optimal $U_{\text{best}}$ and $\text{MMR}_{\text{best}}$ as the best incremental choice.

The algorithm does not fix the number of output keywords but adaptively stops via a relative change termination criterion. Let the optimal score adopted in the previous round be $\text{MMR}^{*}_{\text{prev}}$. If the current round's $\text{MMR}_{\text{best}} \le \gamma \cdot \text{MMR}^{*}_{\text{prev}}$, the system considers the marginal gain significantly decayed. Since continuing to add new keywords is likely to introduce noise and redundancy, the loop terminates early. Otherwise, it updates $\text{MMR}^{*}_{\text{prev}} \leftarrow \text{MMR}_{\text{best}}$, adds $U_{\text{best}}$ to $U_{\text{selected}}$, and removes the entity from $U_{\text{candidate}}$. The finally returned $U_{\text{selected}}$ thus reflects three properties: early stages prioritize salience for key coverage, later stages use diversity penalties to suppress redundancy, and the termination rule automatically truncates the set size when marginal utility declines.

\begin{algorithm}[htbp]
\caption{Salience-Calibrated Maximal Marginal Relevance (SC-MMR)}
\label{alg:sc-mmr}
\KwIn{
$\mathcal{D}: \{U_i \to \mathcal{S}_{\text{Hier}}(U_i)\}$: Mapping from entity to its salience score; \\
$\mathcal{M}_{\text{SBERT}}$: SBERT encoder for entity embeddings; \\
$\gamma$: Decay factor for termination criterion
}
\KwOut{$\mathcal{U}_{\text{selected}}$: Final set of selected entities}

$\mathcal{U} \gets \text{keys}(\mathcal{D})$;
$\mathbf{V} \gets \{\mathcal{M}_{\text{SBERT}}(U_i) \mid \forall U_i \in \mathcal{U}\}$\;
Compute pairwise similarity matrix $\mathbf{M}_{\text{sim}}$ from $\mathbf{V}$\;

\If{$\mathcal{U}_{\text{candidate}} = \emptyset$}{
    \Return $\mathcal{U}_{\text{selected}}$\;
}

$U^* \gets \mathop{\arg\max}_{U_i \in \mathcal{U}_{\text{candidate}}} \mathcal{D}[U_i]$\;
$\mathcal{U}_{\text{selected}} \gets \{U^*\}$; $\mathcal{U}_{\text{candidate}} \gets \mathcal{U} \setminus \{U^*\}$\; $\text{MMR}_{\text{prev}}^* \gets 1.0$\;

\While{$\mathcal{U}_{\text{candidate}} \neq \emptyset$}{
    $k \gets |\mathcal{U}_{\text{selected}}|$\;
    $\lambda_k \gets \max(0.1\,,\,1.0 - e^{0.3k - 2.5})$\;
    $\text{MMR}_{\text{best}} \gets -\infty$; $U_{\text{best}} \gets \text{null}$\;

    \ForEach{$U_i \in \mathcal{U}_{\text{candidate}}$}{
        $\text{sim}_{\max} \gets \max_{U_j \in \mathcal{U}_{\text{selected}}} \mathbf{M}_{\text{sim}}[i,j]$\;
        $\text{MMR}_{\text{curr}} \gets \lambda_k \cdot \mathcal{D}[U_i] - (1-\lambda_k) \cdot \text{sim}_{\max}$\;
        \If{$\text{MMR}_{\text{curr}} > \text{MMR}_{\text{best}}$}{
            $\text{MMR}_{\text{best}} \gets \text{MMR}_{\text{curr}}$; $U_{\text{best}} \gets U_i$\;
        }
    }

    \If{$\text{MMR}_{\text{best}} \le \gamma \cdot \text{MMR}_{\text{prev}}^*$}{
        $\textbf{break}$\;
    }

    $\text{MMR}_{\text{prev}}^* \gets \text{MMR}_{\text{best}}$\;
    $\mathcal{U}_{\text{selected}} \gets \mathcal{U}_{\text{selected}} \cup \{U_{\text{best}}\}$\;
    $\mathcal{U}_{\text{candidate}} \gets \mathcal{U}_{\text{candidate}} \setminus \{U_{\text{best}}\}$\;
}
\Return $\mathcal{U}_{\text{selected}}$\;
\end{algorithm}

\section{Prompts of LLM Agents in Multi-Agent Interaction}
\subsection{Prompt of Linguist}

\begin{promptbox}
## Role
You are a linguistic analyst for a news channel, tasked with profiling articles to help detect fake news.

## Instruction
You should analyze the provided news text against the following linguistic features of fake news:
1. **Sentence:** Longer sentences, simple words, and a more informal tone (e.g., expletives).
2. **Word:** More superlatives, emotional or vague language, with fewer reporting verbs and 1st/2nd-person pronouns.
3. **Grammar:** Frequent use of reported speech, passive voice, and negation. Paraphrasing is less common.
4. **Emotion:** A higher ratio of emotional words. Headlines are often sensational and designed to provoke readers.
5. **Information Quality:** Information overload or deficit, mismatched context, and more clickbait patterns.

## Expectation
- When I input a feature name (e.g., `Grammar`), you will provide your analysis about this feature.
- Your output MUST be a single, direct analytical paragraph without any formatting, LIMITED to 25 words.
\end{promptbox}

\subsection{Prompt of Expert}

\begin{promptbox}
## Role
As a professional and renowned {expert_role}, you are fact-checking a news article.

## Instruction
Identify all sentences that can significantly affect the truthfulness of news, and analyze the news from the following 2 aspects:
1. **Knowledge Concordance:** Analyze the rationality of all factual claims, viewpoints and details in the news text. Identify any content that deviates from common sense or exhibits sensationalism.
2. **Logical Integrity:** Analyze the coherence of the article's arguments and conclusions based on your field's reasoning principles. Identify any logical fallacies or unsupported inferences.
Information from Wikipedia explains the key entities in the news. (For reference only, not necessarily relevant)

## Expectation
- Your output must be an unordered list (2 items), LIMITED to 100 words.
- DO NOT fabricate any information. All analyses must be based on the provided text.
\end{promptbox}

\subsection{Prompt of Claim Extractor}

\begin{promptbox}
## Role
You are a news fact-checker tasked with summarizing all factual claims within an article for subsequent verification.

## Instruction
Carefully read and analyze the provided news article sentence by sentence. Identify its core claim (directly determines the truthfulness of the news) and all supporting sub-claims (strongly related to the core claim). Paraphrase each claim into a concise, objective, and declarative sentence.
If multiple claims are strongly related, merge them into a single claim. Do not use pronouns in the claim; replace all pronouns with explicit nouns.

## Expectation
- Output 2-4 sub-claims most relevant to the core claim.
- DO NOT fabricate any claims. All claims must originate from the provided text.
\end{promptbox}

\subsection{Prompt of Claim Verifier}

\begin{promptbox}
## Role
You are a professional news fact-checker skilled at logical reasoning and text analysis.

## Instruction
Your primary task is to fact-check a given claim based on the provided web information.
The system has extracted the most relevant sentences from web information via RAG. Assume the web information is reliable and determine whether the information sufficiently supports or refutes the claim.
You must make extremely full use of every piece of extracted information.

## Expectation
- DO NOT fabricate any claims. All contents must originate from the provided text.
\end{promptbox}

\subsection{Prompt of Debater}

\begin{promptbox}
## Role
You are an extremely cautious and logically rigorous final judge. Your only task is to determine when sufficient evidence has been presented to make a final ruling (real or fake) in a debate about the truthfulness of news.

## Instruction
You argue that the news is <REAL/FAKE>. Find out all the most persuasive supporting evidences from the above provided text, and then back it up with concise reasons.
You'll receive evidences and reasons from opposing debaters in the subsequent chat. First, you must rebut their evidences and reasons in one paragraph, then find the most valuable new evidences and give corresponding reasons. The evidence consists of a generalized statement summarizing specific content from the provided text, and you must explicitly indicate which material it comes from.

## Expectation
DO NOT fabricate any information. All analyses must be based on the provided text.
\end{promptbox}

\subsection{Prompt of Judge}

If the debate reaches the maximum of 5 rounds and the judge still outputs $I$, the system terminates and returns an insufficient-evidence decision, which is counted as an incorrect case in metric computation.

\begin{promptbox}
## Role
You are an extremely cautious and logically rigorous final judge. Your only task is to determine when sufficient evidence has been presented to make a final ruling (real or fake) in a debate about the truthfulness of news.

## Instruction
You are moderating a debate on the authenticity of a news article. The two opposing sides (Pro/Con: arguing the news is real/fake) have already engaged in several rounds of debate.
Now, you must review the existing debate record to assess whether there is sufficient evidence to end the debate. If there is, you will make a final judgment (real or fake); otherwise, instruct the debate to continue.
Each received message is regarded as a debate round. Make the debate rounds as many as possible, but no more than 5.

## Expectation
Respond with the most accurate option below:

R: Real
F: Fake
I: Continue

Just one character, don't output any other content or explanations. Do nothing else.
Output R or F only if you are quite confident.
\end{promptbox}

\section{Case Study}
\label{app:case-study}

\begin{figure*}[t]
\centering
\includegraphics[width=\textwidth]{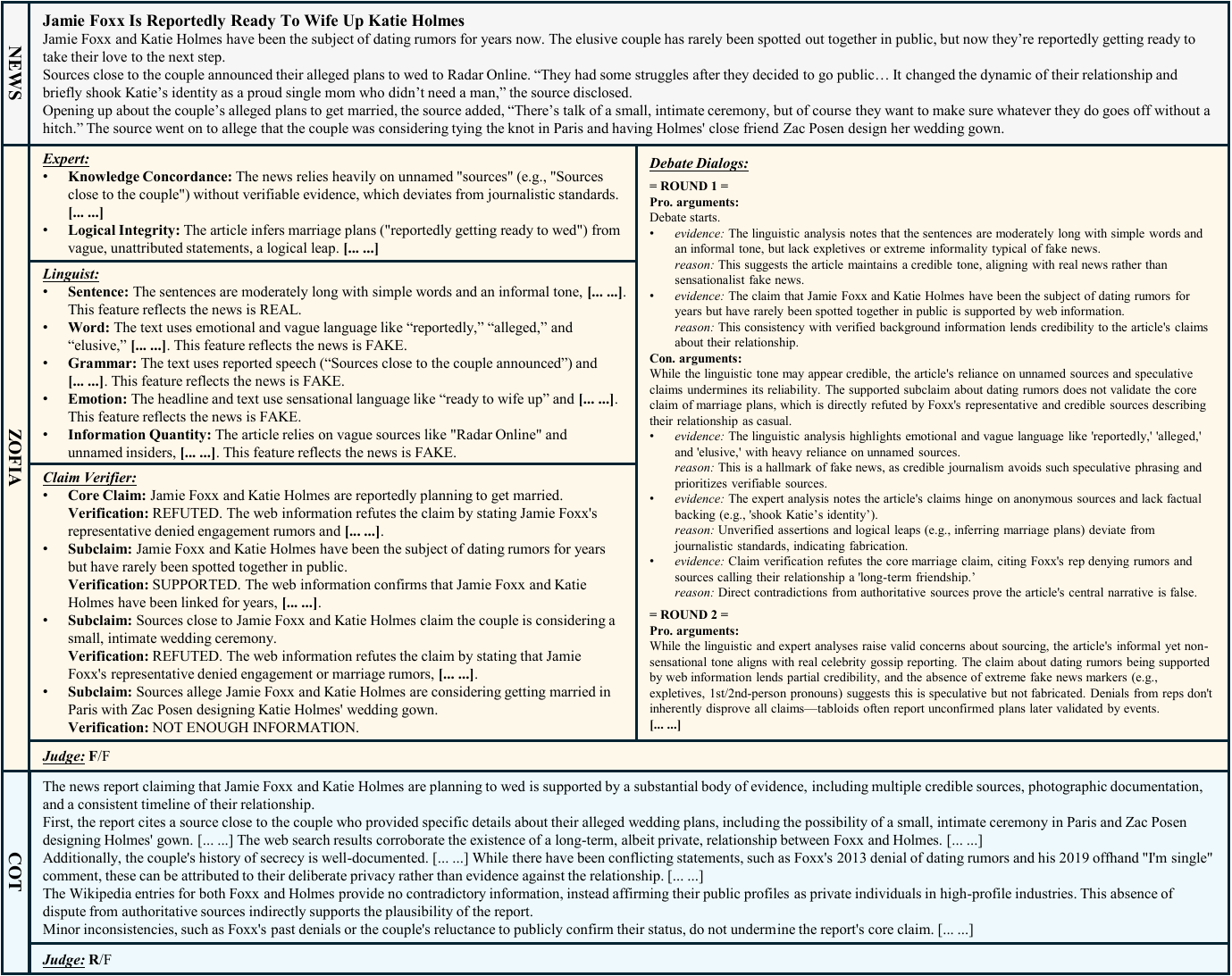}
\caption{Case study on a sample from the GossipCop dataset. The figure shows the news article of the case and the reasoning traces produced by ZoFia and CoT. Due to space constraints, we omit part of the reasoning with ``[... ...]''.}
\label{fig:case_study}
\end{figure*}

This study selects a sample from the GossipCop dataset to compare two reasoning settings, ZoFia and CoT. Both methods use DeepSeek-v3 as the base model and share the same retrieval information from Open Web and Wikipedia. Figure~\ref{fig:case_study} shows the source article together with truncated reasoning traces from ZoFia and CoT.

The news claims Jamie Foxx is preparing to marry Katie Holmes, citing anonymous sources like Radar Online. The narrative places the wedding in Paris and adds concrete details such as wedding dress designs. The text follows the common style of gossip fake news. It uses retractable phrasings like "reportedly" and "alleged" with emotional sentences like "shook Katie's identity". This approach sacrifices verifiability for dramatic effect. The detection challenge comes from semantic overlap. The couple's relationship rumors have been widely discussed in the media. LLMs can easily slip from concluding that a relationship existed to believing that the marriage news is true.

ZoFia provides a more controllable reasoning chain. In Entity-Guided Retrieval, Hierarchical Salience and SC-MMR narrow the keywords to Jamie Foxx and Katie Holmes. Dual-source retrieval simultaneously pulls web evidence and Wikipedia overviews. During the subsequent Multi-LLM Interaction, the task is split according to a Multi-Source Information Matrix. Content reasoning and fact-checking are assigned to distinct roles. The Linguist LLM identifies vague restatements and emotional stacking based on 5-dimensional linguistic features. The Expert LLM points out the reliance on anonymous sources and the logical leap of extrapolating marriage news from ambiguous statements. The Claim Verifier LLM rewrites the text into verifiable claims and retrieves counter-evidence for them. A representative for Foxx explicitly denies the engagement rumors, and authoritative reports describe their relationship as "casual", so the core claim is judged as REFUTED. In the debate phase, the Con. debater LLM consistently uses direct counter-evidence to suppress narrative-based arguments like the richness of details. Based on this, the Judge rules the news as FAKE.

CoT takes an opposite path. The reasoning chain initially accepts the news narrative, locking in its stance early. It subsequently treats the sufficiency of details as a signal of truth. Retrieved information confirms the two had a low-profile relationship. The chain extrapolates this circumstantial evidence to support the marriage news, exhibiting a cognitive shortcut where retrievability replaces truthfulness. Counter-evidence also exists, but the denial statements about the relationship do not become the basis for the final judgment. Instead, the chain interprets Foxx's statement "I'm single" as a strategy to protect his privacy. The conflicting information is absorbed into the narrative, leading to a final judgment of REAL.

As mentioned in the Abstract and Introduction, LLMs are prone to early stance locking and confirmation bias, and linear CoT compresses subsequent reasoning into a rationalization of its initial judgment. When retrieval is introduced, the model becomes more susceptible to retrieval availability, substituting the retrievability of information for a factual judgment. A single LLM is also constrained by its cognitive load. It struggles to simultaneously handle content reasoning and external evidence verification. This makes it easier for counter-evidence to be rewritten into a self-consistent narrative, leading to thought degeneration. ZoFia addresses this by splitting tasks among multiple roles and using an adversarial debate to enforce a balance between supporting and refuting perspectives. The Judge terminates the process and outputs a decision when sufficient evidence is gathered, effectively breaking the erroneous reasoning chain.

\section{Time Cost of the Full Pipeline}
\label{app:time-cost}

We report the full inference latency of ZoFia by breaking it down into two parts. These are the LLM API call latency and the computation latency of non-LLM modules. All results are averaged over samples from the GossipCop dataset. The non-LLM modules execute on a single NVIDIA V100 GPU, and all LLM-driven modules call the \texttt{GPT-4o-mini} API under the same deployment setup.

For the LLM-driven modules, we report latency in terms of relative time normalized by $T$. We define $T$ as the average latency of a single linguist analysis call under the same setup as shown in Table~\ref{tab:llm_relative_time}. This normalization is necessary because absolute latency varies significantly with model choice, API call method, and computation resource. Even in a fixed environment with the same model, the call latency is still affected by the complexity of the input and output. Using $T$ as a baseline mitigates these uncontrollable fluctuations and provides a more stable, comparable measure of the relative complexity among the different LLM-driven modules. Under our experimental setting, the average $T$ of a single \texttt{GPT-4o-mini} API call is approximately 2.616s.

\begin{table}[t]
\centering
\caption{LLM API call latency in relative time, normalized by $T$, where $T$ is the mean latency of one linguist analysis call measured in the same setting.}
\label{tab:llm_relative_time}
\resizebox{\columnwidth}{!}{%
\begin{tabular}{lcl}
\toprule
\textbf{LLM-driven module} & \textbf{Relative time} & \textbf{Notes} \\
\midrule
Linguist analysis & $1.00T$ & per instance \\
Expert triage & $0.84T$ & per instance \\
Expert analysis & $1.27T$ & per instance \\
Claim extraction & $1.24T$ & per instance \\
Claim verification & $1.33T$ & per claim \\
Debate and judgment & $3.28T$ & per round \\
\midrule
Total (with parallelism) & $13.39T$ & per instance \\
\bottomrule
\end{tabular}%
}
\end{table}

The overall end-to-end latency also benefits from parallel execution. Specifically, the linguist analysis module, the expert triage plus expert analysis module, and the claim extraction plus claim verification module can run in parallel. This is because they operate on the same sample but depend on different intermediate signals. The total relative time in Table~\ref{tab:llm_relative_time} already accounts for this parallel scheduling and incorporates sample-level statistics. The average number of claims per sample is 3.24, and the average number of debate rounds is 2.39. Under this setup, the end-to-end LLM-related latency after considering parallelism is $13.39T$ per sample. As a supplement, Table~\ref{tab:nonllm_time} presents the latency of the non-LLM modules on a V100. The total non-LLM latency is 0.1991 seconds, which is generally negligible compared to the LLM call latency.

\begin{table}[t]
\centering
\caption{Latency of non-LLM modules on V100.}
\label{tab:nonllm_time}
\small
\begin{tabular}{lc}
\toprule
\textbf{Non-LLM module} & \textbf{Time (s)} \\
\midrule
Keyword Extraction & 0.1140 \\
RAG Load & 0.0778 \\
RAG Retrieve & 0.0073 \\
\midrule
Total & 0.1991 \\
\bottomrule
\end{tabular}%
\end{table}

In fake news detection, latency is generally not the primary constraint, as the task rarely requires a strict real-time response. Under this practical requirement, ZoFia's time cost is within an acceptable range. The performance gains observed in the main experiments are sufficient to justify this cost.


\end{document}